\documentclass[letterpaper]{article}
\usepackage{uai2018}
\usepackage[margin=1in]{geometry}

\usepackage{times}

\usepackage{microtype}
\usepackage{graphicx}
\usepackage{subfigure}
\usepackage{booktabs} 
\usepackage{bm}
\usepackage{amsmath}
\usepackage{amsthm}
\usepackage{natbib}
\usepackage{amssymb}
\usepackage{color}
\usepackage{algorithmic}
\usepackage{algorithm}
\usepackage{multirow}

\usepackage[title]{appendix}

\newcommand{\Dcal}{\mathcal{D}}

\newcommand{\Fcal}{\mathcal{F}}
\newcommand{\Gcal}{\mathcal{G}}

\newcommand{\Lcal}{\mathcal{L}}

\newcommand{\Ncal}{\mathcal{N}}
\newcommand{\Ocal}{\mathcal{O}}

\newcommand{\Xcal}{\mathcal{X}}
\newcommand{\Ycal}{\mathcal{Y}}

\newcommand{\fb}{\mathbf{f}}
\newcommand{\gb}{\mathbf{g}}
\newcommand{\EE}{\mathbb{E}} 
\newcommand{\RR}{\mathfrak{R}} 

\newcommand{\sbr}[1]{\left[#1\right]}

\newcommand{\nbr}[1]{\left\|#1\right\|}

\newtheorem{theorem}{Theorem}

\DeclareMathOperator*{\argmax}{arg\,max}  

\title{Learning Deep Hidden Nonlinear Dynamics from Aggregate Data}

\author{Yisen Wang\textsuperscript{1, 2, 3}\ \  Bo Dai\textsuperscript{1}\ \ Lingkai Kong\textsuperscript{1}\ \ Sarah Monazam Erfani\textsuperscript{4}\ \  James Bailey\textsuperscript{4} \ \ Hongyuan Zha\textsuperscript{1}\\
  \textsuperscript{1}Georgia Tech\ \ \ \ \
  \textsuperscript{2}Tsinghua University \ \ \ \ \
  \textsuperscript{3}Ant Financial Services Group \ \ \ \ \
  \textsuperscript{4}University of Melbourne\\
{\tt\small wangys14@mails.tsinghua.edu.cn, bohr.dai@gmail.com, zha@gatech.edu }
}

\begin{document}

\maketitle

\begin{abstract}

Learning nonlinear dynamics from diffusion data is a challenging problem since the individuals observed may be different at different time points, generally following an aggregate behaviour. Existing work cannot handle the tasks well since they model such dynamics either directly on observations or enforce the availability of complete longitudinal individual-level trajectories. However, in most of the practical applications, these requirements are unrealistic: the evolving dynamics may be too complex to be modeled directly on observations, and individual-level trajectories may not be available due to technical limitations,  experimental costs and/or privacy issues. To address these challenges, we formulate a model of diffusion dynamics as the {\em hidden stochastic process} via the introduction of hidden variables for flexibility,
and learn the hidden dynamics directly on {\em aggregate observations} without any requirement for individual-level trajectories. We propose a dynamic generative model with Wasserstein distance for LEarninG dEep hidden Nonlinear Dynamics (LEGEND) and prove its theoretical guarantees as well. Experiments on a range of synthetic and real-world datasets illustrate that LEGEND has very strong performance compared to state-of-the-art baselines.

\end{abstract}

\section{INTRODUCTION}
{\em Diffusion data} is a widespread form of data that involves spatial or status transitions over time, \emph{e.g.}, Brownian movement in physics, cell differentiation or gene expression in biology, molecular motion in chemistry, bird migration in ecology, traffic flows in transportation, population trends in social sciences and so on. Learning the underlying dynamics which governs the evolution of such data is a fundamental problem. It reveals the nature of the dynamical phenomenon, based on which we can make better future predictions. However, in these areas, complete longitudinal individual-level trajectories (\textit{i.e.}, the tracking of one individual over the entire diffusion process) may often not be available due to technical limitations, experimental costs and/or privacy issues. Rather, one often instead observes a random group of independently sampled individuals from the population, and the observations can contain different individuals at different time points. This is common for catch and release experiments in ecology (\textit{e.g.}, bird migration) where it is difficult to observe a single bird  twice \citep{bartholomew2005review}, and in biological research where a cell may need to be sacrificed in order for an observation on it to be made \citep{banks2004probabilistic}. We refer to observations made in these scenarios as {\em aggregate observations} to differentiate them from the case of individual-level trajectories which provide full information. 

\begin{figure}[!t]
\centering
\includegraphics[width=0.8\linewidth]{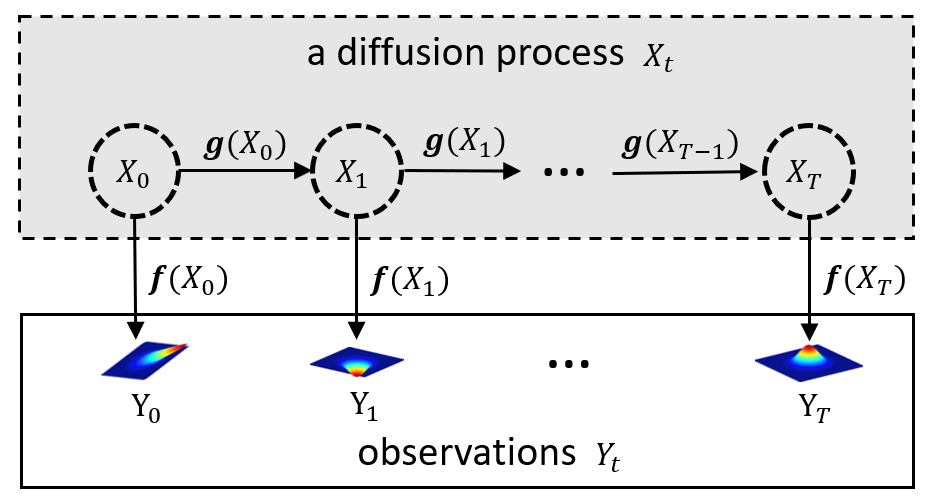}
\caption{An illustration of the framework which builds dynamics on an auxiliary hidden variable $X_t$ with a observation function. Observation $Y_t$ is the aggregate formated data.}
\vspace{-0.1 in}
\label{fig:x-y-diffusion}
\end{figure}

Modeling the dynamics on aggregate observations have been investigated recently in \citep{pmlr-v48-hashimoto16}, where a stochastic differential equation (SDE) has been used to capture the transition directly on observations $Y_t$. However, its performance degrades when the dynamics become complex due to their limited expressive ability, as illustrated later in our experiments. Instead of modeling dynamics directly on observations, a hidden variable $X_t$ can be introduced for modeling complicated dynamics, which can be decomposed into a relatively simple hidden dynamic on $X_t$ with a complicated observation function. As illustrated in Figure \ref{fig:x-y-diffusion}, $Y_t (t \in [0, T])$ is a series of aggregated observations of a diffusion process. We formulate that $Y_t$ is determined by the hidden dynamic on $X_t$ and the observation function $\mathbf{f}(X_t)$. Existing models such as Hidden Markov Model (HMM) \citep{eddy1996hidden}, Kalman Filter (KF) \citep{harvey1990forecasting} and Particle Filter (PF) \citep{djuric2003particle} are popular methods with hidden variables. However, these models and their variants \citep{langford2009learning,hefny2015supervised} require individual-level trajectories, which may not be available, as was mentioned earlier. It consequently still remains an open issue as to how one can learn the underlying dynamics directly from aggregate observations with a hidden stochastic process, for complicated real-world scenarios.

To address these challenges, we propose a novel framework to incorporate the use of hidden variables into the modeling of diffusion dynamics from evolving distributions (as those approximated from aggregate observations). We bypass the need for likelihood-based estimation of model parameters and posterior estimation of hidden variables by using Wasserstein distance learning. The model we propose is named LEGEND (LEarninG dEep hidden Nonlinear Dynamics) and the main contributions are:
\begin{itemize}
\item We propose a framework for learning complicated nonlinear dynamics from aggregate data via a hidden continuous stochastic process.  
\item We extend Wasserstein learning to likelihood-free and posterior-free estimations of dynamic parameter learning and hidden state inference. 
\item We theoretically provide a generalization bound and convergence analysis of our framework, which is the first theoretical result as far as we know.   
\item We empirically demonstrate the effectiveness of our framework for learning nonlinear dynamics from aggregate observations on both synthetic and real-world datasets. 
\end{itemize}

\section{PROBLEM DEFINITION}\label{problem}
We first introduce a continuous model of diffusion dynamics using a stochastic differential equation (SDE), then formally define the tasks of filtering and smoothing based inference, then review the Wasserstein distance objective as one of the distribution metric. 

\textbf{Hidden Continuous Nonlinear Dynamics.} To characterize the dynamics of observations, we introduce a hidden continuous nonlinear dynamical system as shown in Figure \ref{fig:x-y-diffusion}, together with a measurement of hidden states. In particular, the hidden state $X_t \in \mathbb{R}^n$ is the underlying auxiliary variable that cannot be accessed directly, and it follows a SDE:
\begin{equation}\label{def_x}
    dX_t = \mathbf{g}(X_t)dt + \Sigma^{1/2}d\omega_t,
\end{equation}
where $\mathbf{g}:\mathbb{R}^n \rightarrow \mathbb{R}^n$ is a nonlinear deterministic drift function, and $\omega_t \in \mathbb{R}^n$ is a Brownian motion process with noise covariance $\Sigma \in \mathbb{R}^n \times \mathbb{R}^n$. 
At each time point, the observation $Y_t$ is written as a measurement of the hidden state $X_t$:
\begin{equation}\label{def_y}
    Y_t = \mathbf{f}(X_t),
\end{equation}
where $\mathbf{f}:\mathbb{R}^n \rightarrow \mathbb{R}^m$ is a nonlinear observation function. Together, Eqs. (\ref{def_x}) and (\ref{def_y}) define the nonlinear dynamics with continuous hidden states. 

\textbf{Aggregate Observations.} We obtain a collection of independent and identically distributed (i.i.d) samples $\{y_t^i\}_{i=1}^N$ of $Y_t$ at some time point $t$, that we term aggregate or distributional observations. The observed individuals in previous time observations $\{y_{t-1}^i\}_{i=1}^N$ are often not identical to those for the current time observations $\{y_{t}^i\}_{i=1}^N$, implying it is not possible to construct the full trajectory of a single individual. However, we can approximate the probability distribution in terms of a finite number of samples as 
\begin{equation}
    p(Y_t) \approx \frac{1}{N} \sum_{i=1}^{N} \delta(Y_t - y_t^i). 
\end{equation}
Thus, we can treat these aggregate samples from the same time point together as a distribution which evolves in the dynamic system. 

\textbf{Problems.} Under this aggregate setting for observations, it is difficult to obtain individual-level trajectories due to technical limitations, experimental costs and/or privacy issues. Therefore, we propose a new framework to learn the nonlinear dynamics for the distributions (as approximated from aggregate observations) without the need for individual-level trajectories. That is, we treat $\{y_t^i\}_{i=1}^N$ as an empirical approximation to the distribution at time $t$ and its dynamics is learned via an auxiliary hidden variable $X_t$. Once the dynamics are learned, we are faced with two inference tasks: 
\begin{itemize}
    \item[1)] {\em Filtering based inference}: the task is to infer the next future observation $Y_{T+1}$, given observations $\{Y_0, Y_1, \cdots, Y_T\}$. 
    \item[2)] {\em Smoothing based inference}: the task is to infer a missing intermediate observation $Y_k (0< k < T)$, given $\{Y_0, \cdots, Y_{k-1}, Y_{k+1}, \cdots, Y_T\}$. 
\end{itemize}

\textbf{Wasserstein Distance Objective.}
Following our distribution-based problem definition, the metric on distributions is a key criterion for our objection function, just like the mean squared error (MSE) criterion for regression problems. Among the well-known distribution-based measures, such as Total Variation (TV) distance, Kullback-Leibler (KL) divergence, Jensen-Shannon (JS) divergence, Wasserstein distance has recently been shown to possess more appealing properties for distance measurement of distributions \citep{arjovsky2017wasserstein}. We therefore choose to adopt the Wasserstein distance for measuring the discrepancy between the learned distributions and their ground truth.

The definition of Wasserstein-1 distance (also named the Earth-Mover distance (EM)) is:
\begin{equation}
\label{eq:w_define}
    W(\mathbb{P}_r, \mathbb{P}_g) = \inf_{\gamma \in \prod(\mathbb{P}_r, \mathbb{P}_g)} \mathbb{E}_{(x,y) \sim \gamma}[\parallel x-y \parallel], 
\end{equation}
where $\prod(\mathbb{P}_r, \mathbb{P}_g)$ denotes the set of all joint distributions $\gamma(x, y)$ whose marginals are $\mathbb{P}_r$ and $\mathbb{P}_g$ respectively. The infimum in~\eqref{eq:w_define} is highly intractable. However, Kantorovich-Rubinstein duality \citep{villani2008optimal} shows that 
\begin{equation}\label{w_dual}
    W(\mathbb{P}_r, \mathbb{P}_g) = \sup_{\lVert D \rVert_L \le 1} \mathbb{E}_{x \sim \mathbb{P}_r} [D(x)] - \mathbb{E}_{x \sim \mathbb{P}_g} [D(x)], 
\end{equation}
where the supremum is over all $1$-Lipschitz functions $D$. We can assume a parameterized family of functions $\{D_w\}_{w \in \mathcal{W}}$ lying in the $1$-Lipschitz function space. Therefore, Eq.~\eqref{w_dual} could be solved by
\begin{equation}\label{w_solve}
    \max_{w \in \mathcal{W}} \mathbb{E}_{x \sim \mathbb{P}_r} [D_w(x)] - \mathbb{E}_{x \sim \mathbb{P}_g} [D_w(x)]. 
\end{equation}
With Wasserstein distance, our objectives for dynamic learning and inference tasks can be unified to minimize the Wasserstein distance between the generated distribution and the target distribution, which will be instantiated in the following Section 3. 




\section{PROPOSED FRAMEWORK}
In this section, we first discuss our methodology for parameter learning of dynamics within the LEGEND framework, and then introduce in detail how the framework addresses the filtering and smoothing based inference problems.

\subsection{Parameter Learning of Dynamics}
In order to efficiently solve the SDE of hidden state $X_t$, we adopt an approximate numerical solution called the Euler-Maruyama method \citep{talay1994numerical}. Suppose the SDE is defined on $[0, T]$, then the Euler-Maruyama approximation to the true solution of SDE is a Markov chain defined as follows:
\begin{equation}\label{discrete_sde}
    X_{t+\Delta t} = X_t + \mathbf{g}(X_t)\Delta t + \Sigma^{1/2} \Delta \omega_t,
\end{equation}
where the interval $[0, T]$ is partitioned into $M$ equal sub-intervals of width $\Delta t = T/M > 0$ and $\Delta \omega_t$ are independent and identically distributed normal random variables with expected value zero and variance $\Delta t$. Correspondingly, observations $Y_t$ are functions of $X_t$:
\begin{equation}
    Y_t = \mathbf{f}(X_t).
\end{equation}
Given a sequence of distributional observations, we need to minimize the Wasserstein distance between the generated distribution and the observed distribution at each time point to learn functions $\mathbf{f}$ and $\mathbf{g}$. 
The objective function for parameter learning of dynamics is
\begin{equation}\label{l1}
    \min_{\mathbf{f, g}} \sum_t W(\mathbb{P}(Y_t), \mathbb{P}(\mathbf{f}(X_t)),
\end{equation}
where $X_t \sim \mathbb{P}(X_t|X_{t-1})$. The evolving process\footnote{There may be several $\Delta t$ intervals between time $t-1$ and $t$} from $X_{t-1}$ to $X_t$ is controlled by function $\mathbf{g}$ following the SDE in Eq. (\ref{discrete_sde}). One common approach for learning $\mathbf{f}$ and $\mathbf{g}$ is to calculate the likelihood of $Y_t$ under distributions, however in many cases, this is intractable. Here, we propose to use generative models to directly generate samples which satisfy the target distribution $Y_t$. Following the minimization of the Wasserstein distance between generations and observations, the generative model can eventually learn the dynamics of $Y_t$. The specific form of parameterization will be described in Section 4.

\subsection{Filtering based Inference}
The inference of $Y_{T+1}$ given observations $\mathcal{Y}_T = \{Y_0, Y_1, \cdots, Y_T\}$ can be solved by:
\begin{equation}\label{filtering_based_pred}
    Y_{T+1} = (\mathbf{f} \circ \mathbf{g})(X_T). 
\end{equation}
To achieve this, we need to obtain the hidden state $X_T$ first, that is, find the posterior probability $p(X_T|\mathcal{Y}_T)$ of the hidden state conditioned on the entire sequence of observations $\mathcal{Y}_T$, which is a filtering problem. 

To obtain the posterior of the hidden state, the classical forward algorithm needs to solve one dynamic programming problem per sample, which requires individual-level trajectory for posterior inference. However, for our aggregate observation setting, we alternatively treat the Bayesian inference problem from an optimization perspective following \citep{dai2016provable}. 

We first briefly introduce the idea of the optimization method, then generalize it to solve our problem.
\cite{dai2016provable} use a probability $q(U)$ to approximate the posterior probability $p(U|V)$ by minimizing 
\begin{equation}\label{bayes_opt}
    \min_{q(U) \in \mathcal{P}} - \langle q(U), \log p(V|U) \rangle + KL(q(U) \parallel p(U)) 
\end{equation}
over the space of all valid densities $\mathcal{P}$. $\langle \cdot \rangle$ is the inner product, $KL$ is the Kullback-Leibler divergence, $U$ is the hidden variable and $V$ is the observation variable. Thus, $p(V|U)$ is the likelihood of observation and $p(U)$ is the prior of the hidden variable. 
Assuming we have the trajectory for a single individual ($x_t \sim X_t, y_t \sim Y_t$), then the posterior probability of filtering is 
\begin{equation}
    p(x_t|y_{1:t}) = \frac{p(y_t|x_t)p(x_t|y_{1:t-1})}{\int p(y_t|x_t)p(x_t|y_{1:t-1})dx_t},
\end{equation}
where $p(x_t|y_{1:t-1})$ is the propagation probability and $p(x_t|y_{1:t})$ is the updated probability. Generally, $p(x_t|y_{1:t-1})$ could be regarded as the prior of $x_t$ for the updated probability $p(x_t|y_{1:t})$. Following the idea of Eq. (\ref{bayes_opt}), we can use a probability $\pi(x_t)$ to approximate the posterior probability $p(x_t|y_{1:t})$ by recursively optimizing 
\begin{equation}\label{op_1}
\small
    \min_{\pi(x_t) \in \mathcal{P}} - \langle \pi(x_t), \log p(y_t|x_t) \rangle + KL(\pi(x_t) \parallel p(x_t|y_{1:t-1})), 
\end{equation}
where 
\begin{equation}\label{op_2}
\begin{split}
    p(x_t|y_{1:t-1}) & = \int p(x_t, x_{t-1}|y_{1:t-1})dx_{t-1} \\
    & = \int p(x_t|x_{t-1})p(x_{t-1}|y_{1:t-1})dx_{t-1} \\
    & = \int p(x_t|x_{t-1})\pi(x_{t-1})dx_{t-1}.
\end{split}
\end{equation}

In the following, we generalize Eqs. (\ref{op_1}) and (\ref{op_2}) which were defined on an individual trajectory, to the case for aggregate/distributional data. In Eq. (\ref{op_1}), to obtain the optimal solution, we need to maximize the inner product (the first term) and minimize the KL divergence (the second term). We redefine the two terms using Wasserstein distance. For the first term, since maximizing the inner product is equivalent to minimizing the distance between distributions, we replace the inner product with the Wasserstein distance between the distributions of $\mathbf{f}(\pi_t)$ (generated) and $Y_t$ (ground truth). For the second term, we replace KL divergence with Wasserstein distance which is a better measurement for distributions and is thus more suitable for aggregate data \citep{arjovsky2017wasserstein}. In Eq. (\ref{op_2}), the relationship between two consecutive time of hidden variables is replaced by function $\mathbf{g}$. We then can generalize Eqs. (\ref{op_1}) and (\ref{op_2}) to our filtering objective function under aggregate observations: 
\begin{equation}\label{l2}
\begin{split}
     \min \sum_t W(\mathbb{P}(\mathbf{f}(\pi_t)), \mathbb{P}(Y_t)) + W(\mathbb{P}(\pi_t), \mathbb{P}(\mathbf{g}(\pi_{t-1})),
\end{split}    
\end{equation}
where $\pi_t \sim \mathbb{P}(X_t|\mathcal{Y}_t)$ is our target filtering distribution. 




\subsection{Smoothing based Inference}
Smoothing based inference is for predicting the missing intermediate observation $Y_{k} (0< k < T)$ given observations $\mathcal{Y}_{T\setminus k} = \{Y_0, \cdots, Y_{k-1}, Y_{k+1}, \cdots, Y_T\}$. One method \citep{desbouvries2011direct} to solve this is
\begin{equation}\label{smoothing_based_pred}
    Y_{k} = (\mathbf{f} \circ \mathbf{g})(X_{k-1}). 
\end{equation}
To achieve this, we need to obtain the hidden state $X_{k-1}$ first. This is a smoothing problem which focuses on a hidden state somewhere in the middle of a sequence conditioned on the whole sequence of observations. 

Different from the filtering task where the current state is estimated recursively from all past observations, smoothing computes the best state estimates given all available observations from both the past and the future. One well-known and simple approach for smoothing is the forward-backward smoother. During a forward pass the standard filtering algorithm is applied to the observations. Afterwards, on the backward pass, inverse filtering is applied to the same time series of observations. Finally the filtering estimates of both the forward and backward pass are combined into the smoothed estimates \citep{briers2010smoothing}. Since the information from the observation should be incorporated only once into the smoothed estimate, we need to combine the posterior estimate of the forward pass with the prior estimate of the backward pass and vice versa. 

Thus, following the idea in the above filtering problem (treating the posterior estimation from a optimization perspective), we first learn a forward estimate of the hidden state $\pi_{t}^f$ and also a backward estimate $\pi_{t}^b$ using Eq. (\ref{l2}). These then form a weighted Wasserstein barycenter problem \citep{agueh2011barycenters} whose solution is the posterior of smoothing \citep{kitagawa1994two}. That is, we can obtain the optimal smoothing result $\pi_t^s$ by optimizing the Wasserstein distance to the observations and the weighted Wasserstein barycenter:
\begin{equation}\label{l3}
\begin{split}
    & \min \sum_t W(\mathbb{P}(f(\pi^s_t)), \mathbb{P}(Y_t))  \\
    & \qquad + \lambda_1 W(\mathbb{P}(\pi^s_t), \mathbb{P}(\pi_{t}^f)) \\
    & \qquad + \lambda_2 W(\mathbb{P}(\pi^s_t), \mathbb{P}(\pi_{t}^b)),
\end{split}
\end{equation} 
where $\pi_t^s \sim \mathbb{P}(X_t|Y_T)$ is our target smoothing distribution. And the weights $\lambda_1$ and $\lambda_2$ are hyperparameters, which are given intuitively with $\lambda_1 = t/T$ and $\lambda_2 = 1- \lambda_1$ such that smoothing problem becomes filtering problem when $t = T \rightarrow \lambda_1 = 1$. Actually, there are other alternatives one could use for the weights, but these basic settings already work well in our experiments.

\section{MODEL PARAMETERIZATION}
As stated above, we adopt Wasserstein distance to measure the difference between distributions, and have defined our objectives accordingly. In this section, we establish a dynamic generative model via neural network parameterization based on Wasserstein distance.

According to the dual formulation of Wasserstein distance Eq. (\ref{w_solve}), our distribution-based objective of parameter learning of dynamics in Eq. (\ref{l1}) becomes
\begin{equation}\label{l1-gan}
\begin{split}
    & \min_{\mathbf{f,g}} \sum_t \Big( \max_{D_t} \big(\mathbb{E}_{y_t \sim \mathbb{P}(Y_t)}[D_t(y_t)] \\
    & \qquad - \mathbb{E}_{x_t \sim \mathbb{P}(x_t|\mathbf{g}(x_{t-1}))} [D_t(\mathbf{f}(x_t)] \big) \Big).
\end{split}
\end{equation}
For the filtering and smoothing tasks, we introduce a new function $\mathbf{h}$ to characterize the target filtering or smoothing distributions. The filtering objective in Eq. (\ref{l2}) becomes 
\begin{equation}\label{l2-gan}
\begin{split}
     & \min_{\mathbf{h}} \sum_t \Big( \max_{D_t^1} (\mathbb{E}_{y_t \sim \mathbb{P}(Y_t)}[D_t^1(y_t)] \\
     & \qquad - \mathbb{E}_{\pi_t \sim \mathbb{P}(\mathbf{h}(Y_t))} [D_t^1(\mathbf{\mathbf{f}}(\pi_t))]) \\
                & \quad + \max_{D_t^2} (\mathbb{E}_{\pi_t \sim \mathbb{P}(\mathbf{h}(Y_t))} [D_t^2(\pi_t)] \\
                & \qquad - \mathbb{E}_{\pi_{t-1} \sim \mathbb{P}(\mathbf{h}(Y_{t-1}))} [D_t^2(\mathbf{g}(\pi_{t-1}))]) \Big). 
\end{split}    
\end{equation}
And the smoothing objective in Eq. (\ref{l3}) becomes
\begin{equation}\label{l3-gan}
\begin{split}
     & \min_{\mathbf{h}^s} \sum_t \Big( \max_{D_t^1} (\mathbb{E}_{y_t \sim \mathbb{P}(Y_t)}[D_t^1(y_t)] \\
     & \qquad - \mathbb{E}_{\pi_t^s \sim \mathbb{P}(\mathbf{h}^s(Y_t))} [D_t^1(\mathbf{f}(\pi_t^s))]) \\
                & \quad + \lambda_1 \max_{D_t^2} (\mathbb{E}_{\pi_t^s \sim \mathbb{P}(\mathbf{h}^s(Y_t))} [D_t^2(\pi_t^s)] \\
                & \qquad - \mathbb{E}_{\pi_t^f \sim \mathbb{P}(\mathbf{h}^f(Y_t))} [D_t^2(\pi_t^f)]) \\
                & \quad + \lambda_2 \max_{D_t^3} (\mathbb{E}_{\pi_t^s \sim \mathbb{P}(\mathbf{h}^s(Y_t))} [D_t^3(\pi_t^s)] \\
                & \qquad - \mathbb{E}_{\pi_t^b \sim \mathbb{P}(\mathbf{h}^b(Y_t)), } [D_t^3(\pi_t^b)]) \Big).
\end{split}    
\end{equation}
In traditional implicit generative models, given a random variable $z$ with a fixed distribution $p(z)$, we can pass it through a parametric generator $G_\theta$ (typically a neural network) which directly generates samples following a certain distribution $\mathbb{P}_\theta$. Such design is of high flexibility, as by varying the parameters $\theta$ of the neural networks, we can change this distribution to any distribution of interest. While in our framework, we need a dynamic generative model to match distributions at each time step which can be regarded as a combination of several Generative Adversarial Networks (GANs) \citep{goodfellow2014generative}. Specifically, functions $\mathbf{f}, \mathbf{g}, \mathbf{h}$ are all generators (sharing parameters over time) and $D_t$ is a discriminator at time $t$. We formulate functions $\mathbf{f}, \mathbf{g}$ and $D_t$ as normal feed-forward neural networks\footnote{$\mathbf{g}$ could be several nested $\mathbf{g}$ due to the $\Delta t$ in SDE.}:
\begin{align}
    f^l & = \sum_k \sigma(w_k^f f^{l-1} + b_k^f), \\
    g^l & = \sum_k \sigma(w_k^g g^{l-1} + b_k^g), \\
    D_t^l & = \sum_k \sigma(w_k^{D_t} D_t^{l-1} + b_k^{D_t}),
\end{align}
where $f^l, g^l, D_t^l$ are the $l$-th layers of the neural networks, $\sigma$ is the activation function, and $w_k, b_k$ are the neural network parameters. Note the output layer of the discriminator $D_t$ only has one single neuron to output a scalar value.

As for the function $\mathbf{h}$ (in both filtering and smoothing), we use a recurrent neural network (RNN) to model it, similar to \citep{mogren2016c}. For the purposes of simplicity and clarity of exposition, we illustrate the computational process here using a vanilla RNN, whereas the actual recursive unit used in our experiments is LSTM unit \citep{hochreiter1997long}. Given inputs as sequences of observations $\{Y_0, Y_1, \cdots, Y_t, \cdots, Y_T\}$ and outputs as filtering or smoothing hidden states $\{\pi_0, \pi_1, \cdots, \pi_t, \cdots, \pi_T\}$, the parameterization function $\mathbf{h}$ works as follows:
\begin{align}
    s_t & = \sigma(AY_t + Bs_{t-1} + b), \\
    \pi_t & = \sigma(Cs_t+b),
\end{align}
where $s_t$ is the memorized history information and $A, B, C$ are parameter matrices of RNN. 

Note that we need to enforce the Lipschitz constraints when solving Wasserstein distance from duality in Eq. (\ref{w_dual}). To achieve this, we adopt the strategy of gradient penalty in \citep{gulrajani2017improved} to regularize the Wasserstein distance\footnote{We omit this term in our equations for simplicity.}.

For parameter learning of dynamics in Eq. (\ref{l1-gan}), we can obtain the optimal discriminator and gradients by 
\begin{align}\label{eq:grad_estimator}
\small
    D_t^* &= \argmax_{D_t} (E_{y_t^i}[D_t(y_t^i)] - E_{x_{t-1}^i} [D_t((\mathbf{f} \circ \mathbf{g})(x_{t-1}^i))]) \\
    g_{\mathbf{f,g}} &= -\sum_t\nabla_{\mathbf{f,g}} E_{x_{t-1}^i} [D_t^*((\mathbf{f} \circ \mathbf{g})(x_{t-1}^i))],
\end{align}
where the gradient of $\mathbf{g}$ needs to back propagate through the entire chain. In practice, we use gradient decent to update the discriminator. The parameter learning procedure of our model is presented in Algorithm \ref{alg1}. Similarly, we can derive the results for filtering and smoothing from Eq. (\ref{l2-gan}) and (\ref{l3-gan}), respectively.



\begin{algorithm}
\caption{Parameter learning of dynamics}\label{alg1}
\begin{algorithmic}
\FOR{\# training iterations}
    \FOR{$k$ steps}
    \STATE Sample $\{\varepsilon^i\}_{i =1}^N \sim \mathbb{P}(\varepsilon)$ 
        \FOR{time $t$ in [0:T]}
        \STATE Sample $\{y_t^i\}_{i =1}^N \sim \mathbb{P}(Y_t)$
        \FOR{i= 1 to N}
             \STATE $x_0^i = \varepsilon^i$,
             \STATE $x_{t+\Delta t}^i = x_{t}^i + g(x_{t}^i) \Delta t + \Sigma^{1/2} \mathcal{N}(0, \Delta t)$
        \ENDFOR
        \ENDFOR
    \STATE Update the discriminator $D_t$ by $\nabla_{D_t} \frac{1}{N} \sum_{i=1}^N D_t(y_t^i) - \nabla_{D_t} \frac{1}{N} \sum_{i=1}^N D_t((\mathbf{f} \circ \mathbf{g})(x_{t-1}^i))$ 
    \ENDFOR
    \STATE Update $\mathbf{f}, \mathbf{g}$ by ascending its stochastic gradient $ -\sum_t\nabla_{\mathbf{f}, \mathbf{g}} \frac{1}{N} \sum_{i=1}^N D_t((\mathbf{f} \circ \mathbf{g})(x_{t-1}^i))$
\ENDFOR
\end{algorithmic}
\end{algorithm}

\section{THEORETICAL ANALYSIS} 

In this section, we provide a generalization error analysis and a convergence guarantee for our learning framework. Our analysis mainly focuses on the parameter learning component of our method, however, similar results can also be derived for filtering and smoothing based inference. For the purpose of simplicity, we briefly present our main results here and leave detailed theorems and proofs to Appendix~\ref{appendix:sample_complexity}.

\noindent{\bf Generalization Error.} We denote $\Fcal$ and $\Gcal$ as the function spaces of $\mathbf{f}$ and $\mathbf{g}$, respectively, and the $\Dcal$ as the function space of the $\{D_t\}_{t=0}^T$, and $\gb^{\circ t}(x, \xi_t) = \underbrace{((I + \gb)\circ (I + \gb)\circ\ldots\circ (I + \gb))}_{t}(x) + \xi_t$ with $\xi_t\sim \Ncal(0, \Delta t)$. We define
\begin{equation}
\begin{split}
 \ell(\fb, \gb)
= \EE_{y_{0:T}, x_0\sim p(x), \xi_{0:T}}& \bigg[\sum_{t=0}^T \max_{D_t\in \Dcal} \big[D_t(y_t) \\
& - D_t((\fb\circ \gb^{\circ t}(x_0, \xi_{t})))\big] \bigg] .
\end{split}
\end{equation}

\begin{theorem}\label{thm:sample_complexity}
Without loss of generality, we assume in each timestamp the number of the observations is $N$. Given the samples $\Ycal = \{(y^i_t)_{t=0}^T\}_{i=1}^N (|\Ycal|_\infty = C_\Ycal)$ where $y_{0:T} = (y^i_t)_{t=0}^T$ are sampled \emph{i.i.d.} from the underline stochastic processes, and $\Xcal = \{x_0^i\}_{i=1}^N$, $\Xi = \{\xi^i_{0:T}\}_{i=1}^N$ are also \emph{i.i.d.} sampled. Assume $\Dcal$ is a subset of $k$-Lipschitz functions and denote the $\RR(\Fcal\circ\Gcal^{\circ t})$ as the Rademacher complexity of the function space $\Fcal\circ\Gcal^{\circ t}$. We have
\begin{equation}
    \frac{1}{T}\ell(\fb, \gb) \le \frac{1}{T}\hat\ell(\fb, \gb) + \frac{4kC}{\sqrt{N}} + 4k\frac{\sum_{i=1}^T\RR(\Fcal \circ \Gcal^{\circ t})}{T} .
\end{equation}
\end{theorem}

For the different parametrizations, \emph{i.e.}, different function spaces $\Fcal$ and $\Gcal$, the Rademacher complexity of $\RR(\Fcal \circ \Gcal^{\circ t})$ will be different. For example, if we parametrize the $\fb(z) = \sigma(W_fz)$ and $\gb(z, \xi) = I^\top [z, \xi]+ \sigma(W_gz)$ as single layer neural networks, where $\sigma$ satisfies some mild condition~\citep{bartlett2017spectrally}, following~\citep{bartlett2017spectrally}, we have the $\RR(\Fcal \circ \Gcal^{\circ t}) = \tilde\Ocal\bigg(\frac{\sqrt{C_1(W_f)C_2^t(I, W_g)(\sum_{i=1}^tC_3(I, W_g))}}{N}\bigg)$, where $C_1$, $C_2$ and $C_3$ are some constants related to the parameters. For the details of the conditions on $\sigma$ and the exact formulation of the constants, please refer to~\citep{bartlett2017spectrally}.

\noindent{\bf Convergence Analysis.} Convexity-concavity no longer holds for the objectives in the learning and inference parts, therefore, convergence analysis for convex-concave saddle point problem in~\citep{nemirovski2009robust} cannot be directly applied. Inspired by~\citep{dai2017smoothed}, we can see that once we obtain $D_t^*$, Algorithm~\ref{alg1} can be understood as a special case of stochastic gradient descent for a non-convex problem. Thus, we have the following finite-step convergence guarantee for our framework. 

\begin{theorem}\label{thm:convergence}
Assume that the parametrized empirical loss function $\hat\ell(\fb, \gb)$ is $K$-Lipschitz and variance of its stochastic gradient is bounded by $\varsigma^2$. Let the algorithm run $M$ iterations with stepsize $\zeta = \min\{\frac{1}{K}, \frac{C'}{\varsigma\sqrt{M}}\}$ for some $C'>0$ and output $(w_f^1, w_g^1),\ldots, (w_f^M, w_g^M)$. Setting the candidate solution to be $w = (\widehat w_f^M,\widehat w_g^M)$ randomly chosen from $(w_f^1, w_g^1),\ldots, (w_f^M, w_g^M)$ such that $P(w=w^j)=\frac{2\zeta-K\zeta^2}{\sum_{j=1}^N(2\zeta-K\zeta^2)}$, then it holds that
\begin{equation}
    \EE\sbr{\nbr{\nabla \hat\ell(\widehat \fb_w^M,\widehat{\gb}_w^M)}}\leq \frac{K C^2}{M}+ (C'+\frac{C}{C'})\frac{\varsigma}{\sqrt{M}},
\end{equation}
where $C:=\sqrt{2(\hat\ell(w_f^1,w_g^1) -\min \hat\ell(w_f,w_g))/K}$ represents the distance of the initial solution to the optimal solution. 
\end{theorem}

\section{EXPERIMENTS}
In this section, we evaluate our LEGEND framework on various types of synthetic and real-world datasets.

\textbf{Baselines:} We compare our model with two recently proposed methods that learn dynamics directly from aggregate observations --- modeling directly on $Y_t$ using a SDE. The baselines differ from one other in their characterization of the drift term $\mathbf{g}(X_t)$ of the SDE. The two baselines considered in our experiments are two representatives from parametric and non-parametric categories: 1) OU (Orstein-Uhlenbeck \citep{huang2016markov}): modeling the drift term using an Orstein-Uhlenbeck process \citep{gillespie1996exact}, which is a stationary Gauss-Markov process with the drift term $\theta(\mu-x_t)$ ($\theta, \mu$ are parameters); and 2) NN \citep{pmlr-v48-hashimoto16}: modeling the drift term using a neural network (NN) which is a sum of ramps.

\subsection{Synthetic Data}\label{syn_exp}
We first assess our model on three synthetic datasets generated using the following three diffusion dynamics: 




\textbf{Synthetic-1:}
\begin{equation}
\begin{split}
    & x_0 \sim \mathcal{N}(0, \Sigma_1),\\
    & x_{t+\Delta t} = x_t + \frac{1}{4}x_{t} \Delta t  +   \mathcal{N}(0, \Sigma_0), \\
    & y_t = 2x_t.
\end{split}
\end{equation}

\textbf{Synthetic-2:}
\begin{equation}
\begin{split}
    & x_0 \sim \mathcal{N}(0, \Sigma_2),\\
    & x_{t+\Delta t} = x_t + (0.1 x^{2}_{t} +0.5 x_t) \Delta t  + \mathcal{N}(0, \Sigma_0), \\
    & y_t = \exp(x_t).
\end{split}
\end{equation}

\textbf{Synthetic-3:}
\begin{equation}
\begin{split}
    & x_0 \sim \mathcal{U}([-2,2]),\\
    & x_{t+\Delta t} = x_t + (0.5x_{t} + |x_t|) \Delta t  +  \mathcal{N}(0, \Sigma_0), \\
    & y_t = \log |x_t|.
\end{split}
\end{equation}

{\em Synthetic-1} is a linear dynamic on $x_t$ with linear measurement $y_t$ where $x_0$ are sampled from multivariate normal distributions with covariance matrix $\Sigma_1$ (diagonal elements are 0.04 and others are 0.032). A nonlinear dependency between $x_t$ and $y_t$ is formulated in {\em Synthetic-2}: a quadratic dynamic on $x_t$ and an exponential dependency of $y_t$ on $x_t$ where $x_0$ are sampled from multivariate normal distributions with covariance matrix $\Sigma_2$ (diagonal elements are 0.01 and others are 0.008). In {\em Synthetic-3}, we test a more complex scenario: highly nonlinear dynamics on $x_t$ with highly nonlinear measurement $y_t$ where $x_0$ are sampled from a uniform distribution $\mathcal{U}([-2,2])$. For each synthesized dynamic, we obtain $x_t$ like $\{x_0, x_1, x_2, x_3\}$ every $5\Delta t$ time following a discretized SDE in Eq. (\ref{discrete_sde}), and generate 1000 samples at each time step from $x_t$ out of which only 500 samples are chosen as observations $y_t$ like $\{y_0, y_1, y_2, y_3\}$. We consider population evolution $\mathbb{R}^d$ with three different dimensions: $d =2$, $d=5$ and $d=10$. Note that $\Delta t$ is set to $0.2$ for all datasets. The stochastic terms are all sampled from multivariate normal distributions with covariance matrix $\Sigma_0$ (diagonal elements are 0.0025 and others are 0.002). 



Our proposed model along with the baselines are evaluated on two tasks: 1) filtering based inference: given observations $y_0$ and $y_1$, the task is to predict $y_2$; and 2) smoothing based inference: given observations $y_0$, $y_1$ and $y_3$, the task is to predict $y_2$.

\textbf{Experimental Setup:} For our LEGEND model, we set $D$, $\mathbf{f}$, $\mathbf{g}$ as a four-layer, two-layer and four-layer feed-forward neural network respectively with ReLU \citep{glorot2011deep} activation function, and set $\mathbf{h}$ as a one-layer RNN with LSTM unit. In terms of training, we use the Adam optimizer \citep{kingma2014adam} with learning rate $10^{-4}$, $\beta_1 = 0.5$ and $\beta_2 = 0.9$. The baselines OU and NN are configured with respect to their settings in the original papers without using pre-training \citep{pmlr-v48-hashimoto16}.

\begin{figure}[!t]
\centering
\includegraphics[width=0.98\linewidth]{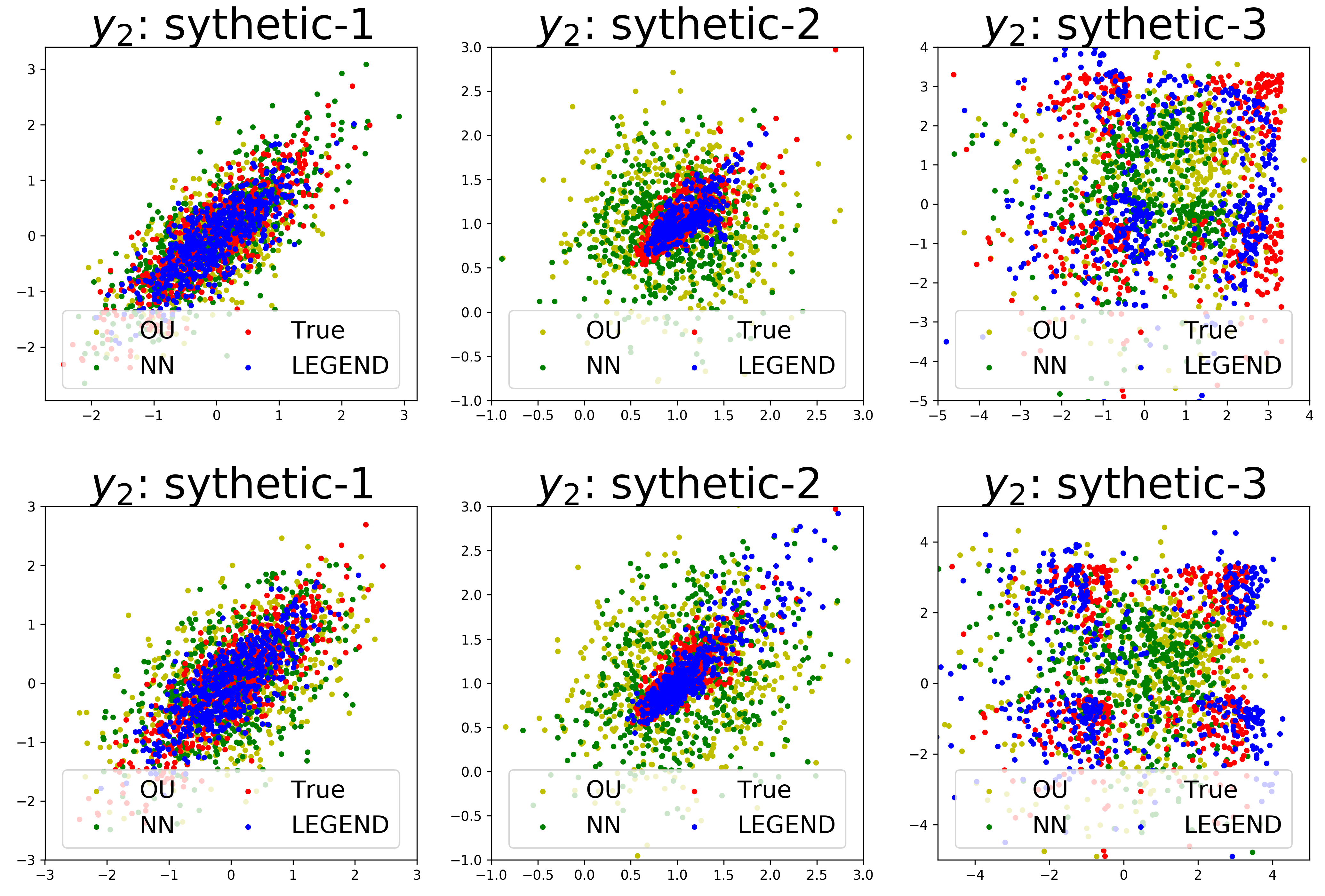}
\caption{The true and predicted distributions for different models in filtering (top row) and smoothing (bottom row) based inference tasks on 2-dimensional synthetic-1 (left column), synthetic-2 (middle column) and synthetic-3 (right column) datasets.}
\label{fig:synthetic_filtering}
\end{figure}

\textbf{Results:} We first show the capability of our model for learning low-dimensional ($d=2$) diffusion dynamics.
As visualized in Figure \ref{fig:synthetic_filtering}, given $\{y_0\}$ and $\{y_1\}$, our model can precisely learn the dynamics and correctly predict $y_2$ (top row) where a better match was observed between the predictions (blue points) and the ground truth (red points). Note that the dynamics on $y_t$ become more and more complicated from synthetic-1 to synthetic-3. It can be seen from Figure \ref{fig:synthetic_filtering} that our model works well on both simple and complex dynamics, while baselines OU and NN only work well on simple dynamics. Similar results are also observed in the smoothing based inference task, as shown in the bottom row of Figure \ref{fig:synthetic_filtering}.

We then evaluate our model using Wasserstein error for both low-dimensional ($d=2$) and high-dimensional ($d=5,10$) diffusion dynamics. Wasserstein error measures the difference between predicted distribution and the true distribution. As reported in Table \ref{tb:werror}, our model achieves much lower Wasserstein error than the two baselines on all the 3 datasets for 2/5/10-dimensional dynamics. The poor performance of OU and NN may due to the fact that $y_t$ becomes more and more complicated as dimension increases on all three datasets. The superior performance of our model verifies the importance of hidden variables --- they are necessary for the modeling of complex nonlinear dynamics and complex measurements of hidden states.

\begin{table}[!tb]
\renewcommand{\arraystretch}{1.1}
\centering
\small
\caption{The Wasserstein error of different models on synthetic-1/2/3 (Syn-1/2/3), RNA-seq (RNA) and bird migration (Bird) datasets. The best results are highlighted in \textbf{bold}.}
\vspace{0.1 in}
\label{tb:werror}
\begin{tabular}{c|c|c|ccc}
\hline
\multirow{2}{*}{Data} & \multirow{2}{*}{Target} & \multirow{2}{*}{Task} & \multirow{2}{*}{NN} & \multirow{2}{*}{OU} & LEG- \\
 &  &  &  &  & END  \\ \hline
\multirow{6}{*}{Syn-1} 
& filtering & $d=2$ & 0.30  & 0.29 & \textbf{0.06}  \\
& $y_2$ & $d=5$ & 3.09  & 2.52 & \textbf{0.06}  \\
& & $d=10$ & 11.19  & 9.61 & \textbf{0.18}  \\
\cline{2-6}
& smoothing & $d=2$ & 0.70  & 0.80 & \textbf{0.04}  \\
& $y_2$ & $d=5$ & 3.40  & 2.92 & \textbf{0.08}  \\
& & $d=10$ & 9.58  & 8.96 & \textbf{0.12}  \\
\hline
\multirow{6}{*}{Syn-2} 
& filtering & $d=2$ & 0.87  & 1.36 & \textbf{0.17}  \\
& $y_2$ & $d=5$ & 3.49  & 4.38 & \textbf{0.47}  \\
& & $d=10$ & 8.55  & 10.42 & \textbf{1.37}  \\
\cline{2-6}
& smoothing & $d=2$ & 1.62  & 1.75 & \textbf{0.22}  \\
& $y_2$ & $d=5$ & 5.28  & 4.17 & \textbf{0.57}  \\
& & $d=10$ & 11.14  & 9.91 & \textbf{2.84}  \\
\hline
\multirow{6}{*}{Syn-3} 
& filtering & $d=2$ &8.55   & 10.79 & \textbf{3.79}  \\
& $y_2$ & $d=5$ & 31.95 &35.17  & \textbf{13.21}  \\
& & $d=10$ & 113.21   & 116.42 &  \textbf{42.52}  \\
\cline{2-6}
& smoothing & $d=2$ &8.43  & 9.08 & \textbf{2.22}  \\
& $y_2$ & $d=5$ &  28.37 & 31.26 & \textbf{11.26}  \\
& & $d=10$ &  102.65 & 109.80 & \textbf{38.73}  \\
\hline \hline
\multirow{4}{*}{RNA} 
& \multirow{2}{*}{Krt8} & D7 & 6.16  & 9.82 & \textbf{2.31}  \\
& & D4 & 27.98  & 24.54 & \textbf{4.89}  \\
\cline{2-6}
& \multirow{2}{*}{Krt18} & D7 & 6.86  & 9.80 & \textbf{3.16}  \\
& & D4 & 24.75  & 25.88 & \textbf{4.21}  \\
\hline \hline
\multirow{2}{*}{Bird} 
& \multirow{2}{*}{GrayJay} & June & 1.9e3  & 2.5e3 & \textbf{1.2e3}  \\
& & April & 1.5e3  & 1.1e3 & \textbf{0.3e3}  \\
\hline 
\end{tabular}
\end{table}


\subsection{Real Data: Single-cell RNA-seq}\label{rna}
In this section, we evaluate our model on a typical application of distribution based continuous diffusion dynamics in biology: learning the diffusion process where embryonic stem cells differentiate into mature cells \citep{klein2015droplet}. The cell population begins to differentiate from embryonic stem cells after the removal of LIF (leukemia inhibitory factor) at day 0 (D0). Single-cell RNA-seq measurements (or observations) are sampled at day 0 (D0), day 2 (D2), day 4 (D4), and day 7 (D7). At each time point, the expression of 24,175 genes for several hundreds cells are measured (933, 303, 683 and 798 cells at D0, D2, D4, and D7 respectively). We focus on the dynamics of cell differentiation for the two main epithelial makers studied in \citep{klein2015droplet}, \textit{i.e.}, Keratin 8 (Krt8) and Keratin 18 (Krt18).
We evaluate two tasks on this data: 1) filtering based inference: predicting the gene expression level at D7 given only the observations at D0 and D4; and 2) smoothing based inference: predicting the gene expression level at D4 given D0, D2 and D7.


\textbf{Experimental Setup:}
We set $\mathbf{f}$ as a one-layer feed-forward neural network, $\mathbf{g}$ as a three-layer feed-forward neural network and $\mathbf{h}$ as a one-layer RNN with LSTM neurons. For preprocessing, we apply standard normalization procedures \citep{hicks2015widespread} to correct for batch effects, and impute missing expression levels using non-negative matrix factorization, similarly as it did in \citep{pmlr-v48-hashimoto16}. The stochastic term $\Sigma$ in SDE are sampled from multivariate normal distributions with diagonal covariance matrix (diagonal elements are 1). 
Other configurations and baselines are the same as those in Section \ref{syn_exp}. 

\begin{figure}[!t]
\centering
\includegraphics[width=0.98\linewidth]{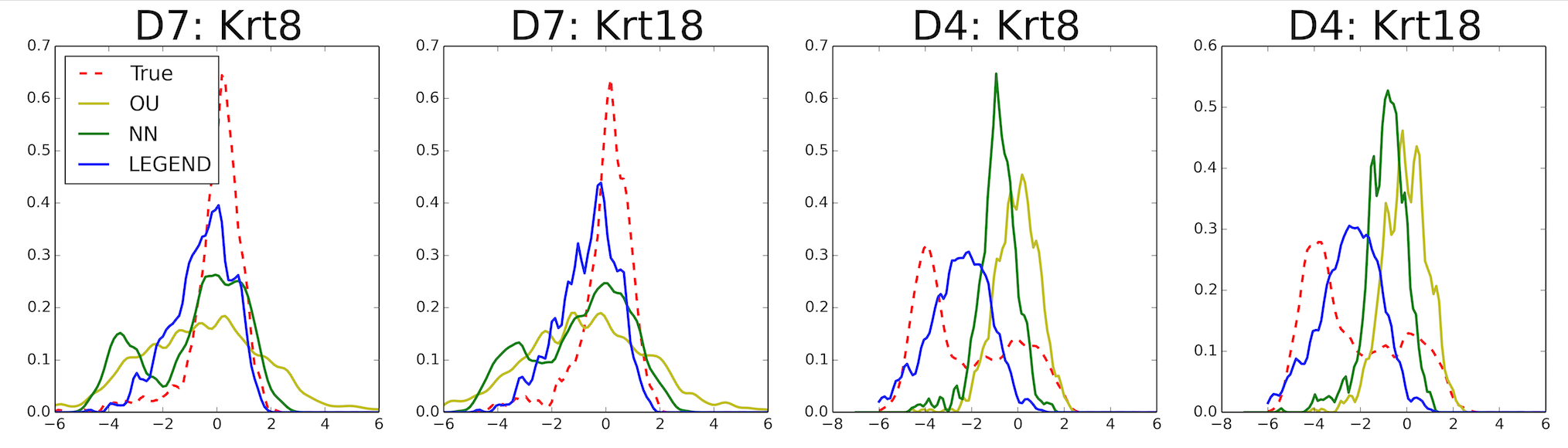}
\caption{The true and predicted marginal distributions of the differentiating genes at D7 (filtering based inference task) and D4 (smoothing based inference task) for different models.}
\label{fig:marginal}
\vspace{0.2 in}
\end{figure}

\begin{figure}[!t]
\centering
\includegraphics[width=0.98\linewidth]{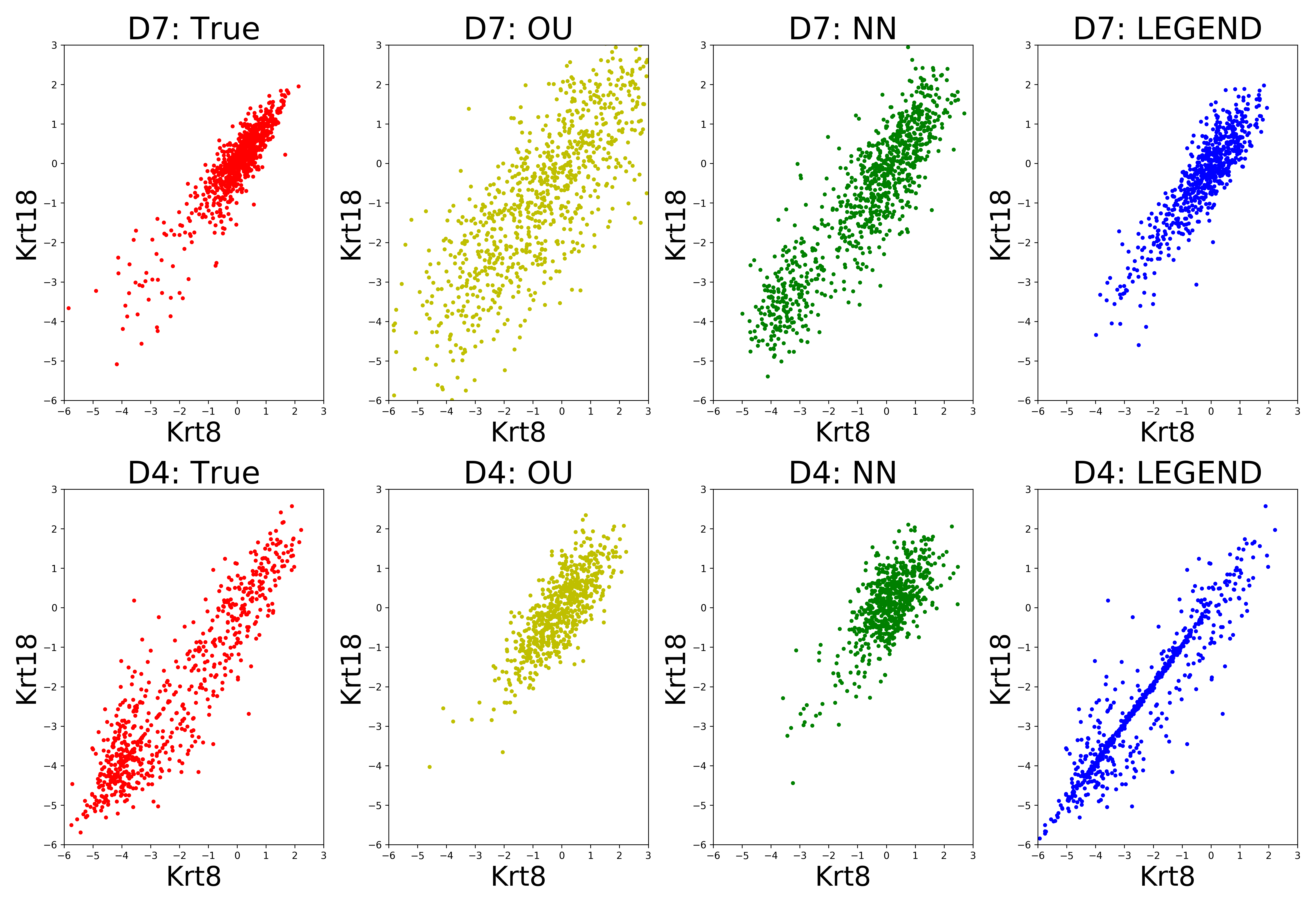}
\caption{The true (left column) and predicted (right 3 columns) correlations between Krt8 and Krt18 at D7 (top row) and D4 (bottom row). The closer to the true correlation the better the performance.}
\label{fig:pairwise}
\end{figure}

\textbf{Results: } 
We first show in Table \ref{tb:werror} that compared to other baselines our model achieves the lowest Wasserstein error in both filtering (D7) and smoothing (D4) tasks on both Krt8 and Krt18. This proves that our model is capable of learning the precise differentiation dynamics and the distributions of the two studied gene expressions. We further provide a closer look into the learned distributions of the two genes in Figure \ref{fig:marginal}. As can be seen, the distributions of Krt8 and Krt18 predicted by our model (curves in blue) are much closer to their true distributions (curves in red) at both D4 and D7, as compared to the baseline models. Moreover, our model can effectively identify the correlations between Krt8 and Krt18, as shown in Figure \ref{fig:pairwise}. This implies that our model can accurately learn the dynamics even considering the correlational structure of the true dynamics. 





\subsection{Real Data: Bird Migration}
We also evaluate our model on another typical application of distribution based diffusion dynamics: bird migration research in ecology. We use the eBird basic dataset (EBD), which gathers large volumes of information on where and when birds occur in the world \citep{sullivan2009ebird}. We down-sampled EBD to only include the tracking records for the species GrayJay between January, 2017 and June, 2017 (monthly data) at United States where 400 samples are randomly selected as observations for each month. There are again two tasks evaluated here: 1) filtering based inference: we apply our model on the months February and April so as to predict the population at June; 2) smoothing based inference: we apply our model on months February, March and June so as to predict the population at April. The experimental setups are the same as those in Section \ref{rna}. 

\begin{figure}[!t]
\centering
\includegraphics[width=0.98\linewidth]{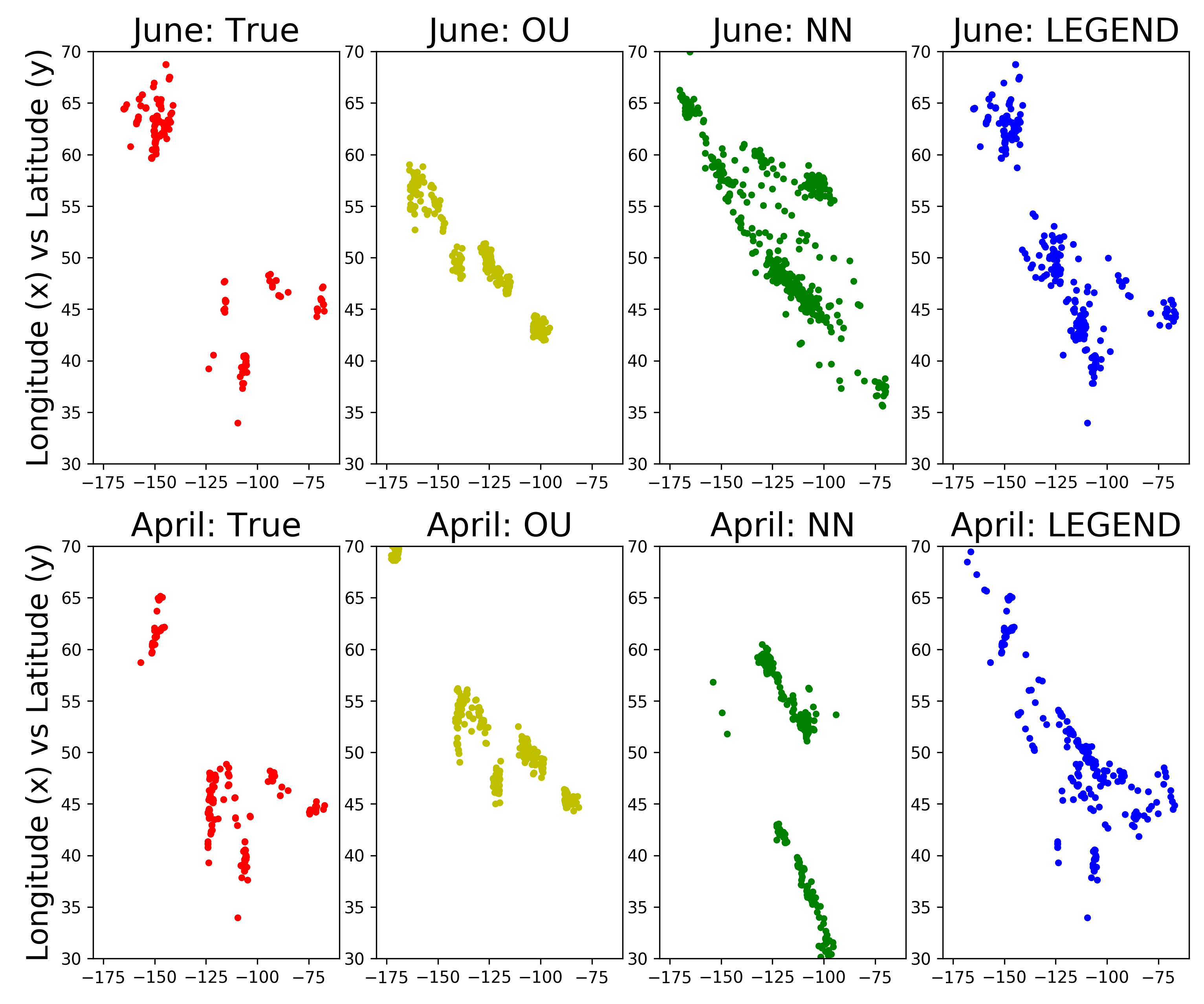}
\caption{The true (left column) and predicted (right 3 columns) distributions of GrayJay species at month June (filtering based inference task) and April (smoothing based inference task) for different models.}
\label{fig:marginal_bird}
\end{figure}

\textbf{Results: } 
We plot the true and predicted locations (longitude and latitude) of the species GrayJay in Figure \ref{fig:marginal_bird}, and report the Wasserstein error\footnote{Our model could be further improved if considering more complex hidden diffusion process, \textit{e.g.}, jump diffusion process, but the framework is the same to this paper.} in Table \ref{tb:werror}. Again, our model achieves the lowest Wasserstein error in both filtering (June) and smoothing (April) based inference tasks. The evolving dynamics of bird migration can be very complicated and extremely difficult to learn, mostly because bird migration could be affected by many irregular factors related to the specific time. Even so, with the introduction of the hidden state variable, our model can predict locations which are close to the ground truth, and with better performance than OU and NN which directly build models on observations. This result demonstrates the advantages of our framework in solving real-world problems involving complicated diffusion dynamics. 

\section{CONCLUSIONS}
In this paper, we formulated a novel technique to learn nonlinear continuous diffusion dynamics from {\em aggregate observations}. In particular, we showed how one can model dynamics as a {\em hidden continuous stochastic process}, and proposed a  framework that employs a dynamic generative model with Wasserstein distance to learn the evolving dynamics. In addition to deriving solutions for both filtering and smoothing based inference tasks, we also established theoretical guarantees on the generalization and convergence properties of our framework. Through comprehensive experimental evaluation on synthetic and real-world datasets, we demonstrated that our approach has very strong performance compared to state-of-the-art techniques on both filtering and smoothing based inference tasks.

\section*{Acknowledgements}
This work is partially supported by NSF IIS-1717916, CIMM-1745382 and China Scholarship Council (CSC).

{
\bibliography{UAI2018_GAN}
\bibliographystyle{icml2018.bst}
}


\vfill\pagebreak
\newpage
\onecolumn 

\begin{appendices}

\begin{center}
{\Large \bf Appendix}
\end{center}

\section{Proof Details of the Theoretical Analysis}\label{appendix:sample_complexity}

\subsection{Generalization Error}
In this section, we analyze the generalization error on the model learning task. We denote $\Fcal$ and $\Gcal$ as the function spaces of $\mathbf{f}$ and $\mathbf{g}$, respectively, and the $\Dcal$ as the function space of the $\{D_t\}_{t=0}^T$, where $T$ stands for the number of steps, and $\gb^{\circ t}(x, \xi_t) = \underbrace{((I + \gb)\circ (I + \gb)\circ\ldots\circ (I + \gb))}_{t}(x) + \xi_t$ with $\xi_t\sim \Ncal(0, \Delta t)$. We define
\begin{equation*}
\ell(\fb, \gb) = \EE_{y_{0:T}, x_0\sim p(x), \xi_{0:T}}\bigg[\sum_{t=0}^T \max_{D_t\in \Dcal} \big[D_t(y_t) - D_t((\fb\circ \gb^{\circ t}(x_0, \xi_{t})))\big] \bigg] := \ell_t(\fb, \gb),
\end{equation*}
where 
$$
\ell_t(\fb, \gb) = \EE_{y_t, x_0, \xi_{0:T}}\bigg[\max_{D_t\in \Dcal} \underbrace{\big[D_t(y_t) - D_t((f\circ g^{\circ t}(x_0, \xi_{t})))\big]}_{\phi_t(\fb, \gb, D_t)} \bigg].
$$

Without the loss of generality, we assume in each timestamp the number of the observations is $N$. Given the samples $\Ycal = \{(y^i_t)_{t=0}^T\}_{i=1}^N$, where $y_{0:T} = (y^i_t)_{t=0}^T$ are sampled \emph{i.i.d.} from the underline stochastic processes, and $\Xcal = \{x_0^i\}_{i=1}^N$, $\Xi = \{\xi^i_{0:T}\}_{i=1}^N$ are also \emph{i.i.d.} sampled, we have the empirical loss function as 
\begin{equation*}
\hat\ell(\fb, \gb) = \hat\EE_{\Ycal}\hat\EE_{\Xcal}\bigg[\sum_{t=0}^T \max_{D_t\in \Dcal} \big[D_t(y_t) - D_t((\fb\circ \gb^{\circ t}(x_0,\xi_{t})))\big] \bigg] = \sum_{t=0}^T \hat\ell_t(\fb, \gb).
\end{equation*}
With the notations defined above, we provide the proof for Theorem~\ref{thm:sample_complexity} as below. 

\begin{proof}
Denote the $\hat \fb$ and $\hat \gb$ are the solutions provided by the algorithm, and $\fb^*$ and $\gb^*$ be the optimal solutions, we have
\begin{eqnarray*}
    |\ell_t(\hat\fb, \hat\gb) - \ell_t(\fb^*, \gb^*)|  &=& \bigg|\EE[\max_{D_t\in \Dcal} \phi_t(\hat\fb,\hat\gb,D_t)] - \EE[\max_{D_t\in \Dcal} \phi_t(\fb^*,\gb^*,D_t)]\bigg| \\
    & \le& \bigg|\max_{D_t\in\Dcal} \EE[\phi_t(\hat\fb,\hat\gb,D_t) - \phi_t(\fb^*,\gb^*,D_t)]\bigg| \\
    & \le& 2 \sup_{\fb\in\Fcal,\gb\in\Gcal,D_t\in\Dcal} |\hat{\Phi}_t (\hat{\fb},\hat{\gb},D_t) - \Phi_t(\fb^*,\gb^*,D_t)|,
\end{eqnarray*}
where
\begin{eqnarray*}
    \hat{\Phi}_t (\hat{\fb},\hat{\gb},D_t) &=& \hat\EE_{y_t\in \Ycal_t}\hat\EE_{x_0, \xi_t}\big[\phi_t(\hat{\fb},\hat{\gb},D_t)\big], \\
    \Phi_t(\fb^*,\gb^*,D_t) &=& \EE\big[\phi_t(\fb^*,\gb^*,D_t)\big].
\end{eqnarray*}
Assume $\Dcal\in\Lcal_k$, where $\Lcal_k$ denotes the $k$-Lipschitz function space, and $|\Ycal|_\infty = C_\Ycal$, we have,
\begin{eqnarray*}
    &&\sup_{\fb\in\Fcal,\gb\in\Gcal,D_t\in\Dcal} |\hat{\Phi}_t (\hat{\fb},\hat{\gb},D_t) - \Phi_t(\fb^*,\gb^*,D_t)| \le 2 \EE\bigg[\sup_{\fb\in\Fcal,\gb\in\Gcal,D_t\in\Dcal} \bigg|\frac{1}{N}\sum_{i=1}^N \tau_i \phi_t (\fb, \gb, D_t)\bigg|\bigg]\\
    &\le& 2 \EE\bigg[\sup_{D_t\in\Dcal} \bigg|\frac{1}{N}\sum_{i=1}^N \tau_i D_t(y_i)\bigg|\bigg] + 2 \EE\bigg[\sup_{\fb\in\Fcal,\gb\in\Gcal,D_t\in\Dcal} \bigg|\frac{1}{N}\sum_{i=1}^N \tau_i  D_t((\fb\circ \gb^{\circ t}(x_0,\xi_{t})))\bigg|\bigg]\\
    &\le& 2\frac{kC}{\sqrt{N}} + 2k\EE\bigg| \frac{1}{N}\sum_{i=1}^N \tau_i  \fb\circ \gb^{\circ t}(x_0,\xi_{t})) \bigg| = 2\frac{kC}{\sqrt{N}} + 2k \RR(\Fcal\circ\Gcal^{\circ t}),
\end{eqnarray*}
where the $\RR(\Fcal\circ\Gcal^{\circ t})$ denotes the Rademacher complexity of the function space $\Fcal\circ\Gcal^{\circ t}$. Therefore, we have
$$
\frac{1}{T}\ell(\fb, \gb) \le \frac{1}{T}\hat\ell(\fb, \gb) + \frac{4kC}{\sqrt{N}} + 4\frac{k\sum_{i=1}^T \RR(\Fcal \circ \Gcal^{\circ t})}{T}.
$$
\end{proof}


\subsection{Convergence Analysis}

Inspired by~\citep{dai2017smoothed}, we can see that once we obtain the $D_t^*$, the Algorithm~\ref{alg1} can be understood as a special case of stochastic gradient descent for non-convex problem. We prove the Theorem~\ref{thm:convergence} as below.
\begin{proof}
We compute the gradient of $\ell(\fb, \gb)$ w.r.t. $\fb$, the same argument is also for gradient w.r.t. $gb$.
\begin{eqnarray}
    \nabla_\fb \ell(\fb, \gb) &=& \nabla_\fb\EE\bigg[\sum_{t=0}^T \phi_t(\fb, \gb, D_t^*)\bigg] = \EE\bigg[\sum_{t=0}^T \nabla_\fb \phi_t(\fb, \gb, D_t^*)\bigg] \\
    &=& \EE\bigg[\sum_{t=0}^T (\nabla_\fb \phi_t(\fb, \gb, D_t^*) + \underbrace{\nabla_{D_t^*} \phi_t(\fb, \gb, D_t^*)\nabla_\fb D_t^*(\fb\circ\gb^{\circ t})}_{0})\bigg]\\
    &=& -\EE\bigg[\sum_{t=0}^T \nabla_\fb D^*_t(\fb\circ\gb^{\circ t})\bigg]
\end{eqnarray}
The second term in the last second line is zero due to the optimality of $D^*$. Therefore, we achieve the unbiasedness of the gradient estimators.  

As long as the gradient estimator for $\fb$ and $\gb$ are unbiased, the convergence rate in Theorem~\ref{thm:convergence} will be automatically hold from~\citep{ghadimi2013stochastic}. 
\end{proof}

\end{appendices}

\end{document}